\documentclass{article}
\usepackage{arxiv}
\usepackage{cite}
\usepackage[utf8x]{inputenc} 
\usepackage[T1]{fontenc}    
\usepackage{hyperref}       
\usepackage{url}  
\usepackage{graphicx}
\usepackage{xcolor}
\usepackage{float}
\usepackage{booktabs}       
\usepackage{amsfonts}       
\usepackage{nicefrac}       
\usepackage{microtype}      
\usepackage{lipsum}
\usepackage{caption}
\usepackage{authblk}
\usepackage[bottom]{footmisc}
\usepackage{adjustbox}
\usepackage{xcolor}
\title{Vanishing Twin GAN: How training a weak Generative Adversarial Network can improve semi-supervised image classification}

\author[1,2,*]{Saman Motamed}
\author[1,2,3]{Farzad Khalvati}
\affil[1]{Institute of Medical Science, University of Toronto}
\affil[2]{Department of Diagnostic Imaging, Neurosciences and Mental Health, The Hospital for Sick Children}
\affil[3]{Department of Mechanical and Industrial Engineering, University of Toronto}
\affil[*]{sam.motamed@mail.utoronto.ca}

\begin{document}
\maketitle
\begin{abstract}
Generative Adversarial Networks can learn the mapping of random noise to realistic images in a semi-supervised framework. This mapping ability can be used for semi-supervised image classification to detect images of an unknown class where there is no training data to be used for supervised classification. However, if the unknown class shares similar characteristics to the known class(es), GANs can learn to generalize and generate images that look like both classes. This generalization ability can hinder the classification performance. In this work, we propose the Vanishing Twin GAN. By training a weak GAN and using its generated output image parallel to the regular GAN, the Vanishing Twin training improves semi-supervised image classification where image similarity can hurt classification tasks.  
\end{abstract}

\keywords{Generative Adversarial Networks \and Image Classification \and Semi-supervised Classification}
\section{Introduction}
Generative Adversarial Networks \cite{goodfellow2014generative} is one of the most exciting inventions in machine learning in the past decade, where the network learns to generate never-before-seen images from a domain it was trained on. Schlegl \emph{et al.}~\cite{schlegl2017unsupervised} used GANs to classify images of retinal fluid or hyper-reflective foci in optical coherence tomography (OCT) images of the retina. By defining a variation score \(V(x)\) (eq. \ref{eq:ax}), their proposed Anomaly Detection GAN (AnoGAN) captured the characteristic and visual differences of two images; one generated by the GAN and one real test image. The idea was to, for instance, train the GAN on only healthy images. When GAN is trained, the generator can generate images similar to those in the healthy image class. During the test phase, the variation score \(V(x)\) must be low if the test image x is healthy and GAN's generator (\textbf{G}) can generate a similar image to that of the healthy image. If the test image is not healthy and varies from the healthy class, \(V(x)\) would be larger, and the generated image would look visually different than the real test image with characteristics that make it non-healthy.
Recently, Deep Support Vector Data Description (Deep SVDD) \cite{ruff2018deep} was proposed that outperformed AnoGAN and shallow models such as Isolation Forest (IF) \cite{liu2008isolation} and OC-SVM \cite{scholkopf2001estimating} in the one class classification framework. Deep SVDD learns a neural network transformation from inputs into a hypersphere characterized by center \emph{c} and radius \emph{R} of minimum volume. The idea is that this allows for the known (normal) class of images to fall into the hypersphere and the unknown (abnormal) class to fall outside of the hypersphere.

\begin{figure}
  \centering
  \includegraphics[scale=0.4]{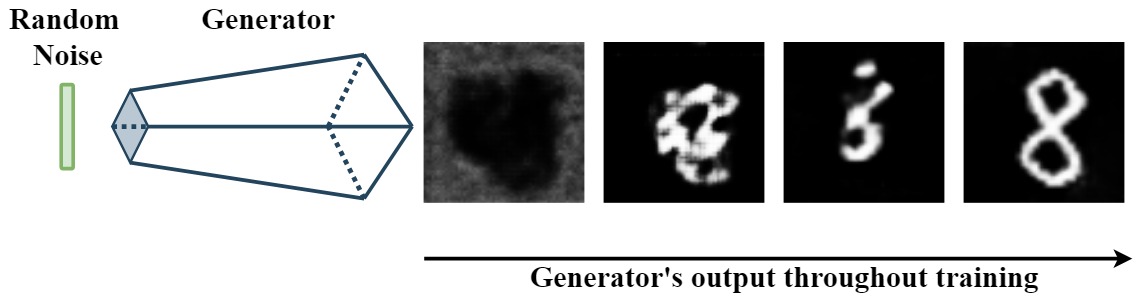}
  \caption{GAN generalization}
  \label{fig:process}
\end{figure}

We observed that GANs when used for classification tasks, can suffer from the generator's generalization ability. Figure \ref{fig:process} shows the training process of a DCGAN \cite{radford2015unsupervised} trained on the MNIST dataset to generate images of handwritten digit 8. The training process takes the 1D random input noise to the 2D image domain. As the training progresses, the figure shows that a noisy image starts to look like the class we want to generate images of (digit 8). However, it can be seen that the second to last image in figure \ref{fig:process} could also be classified as the digit 3. In fact, classes 3 and 8 are similar pairs of classes in the MNIST dataset (figure \ref{fig:2}). 

Recently, Multi-class GAN (MCGAN) \cite{motamed2021multi}, a Generative Adversarial Network for semi-supervised multi-class image classification, attempted to improve this flaw of GANs for image classification. MCGAN used labels from similar classes to enforce the generator not to generalize between similar classes. For instance, in a three-class classification where classes C1 and C2 have labels, and an unknown class C3 does not have labels, MCGAN showed improvement in classifying the classes where Traditional GAN (A traditional GAN, where the discriminator takes two images as an input such as DCGAN) misclassified images of C1 and C2 due to similarity of images between the two classes. MCGAN, however, requires labels from the known classes in order to force non-generalizability. For a semi-supervised binary classification between C1 and C2, MCGAN would fail to improve the results since it does not have access to both class' labels. In this work, we propose Vanishing Twin \cite{landy1998vanishing} GAN (VTGAN). VTGAN improved semi-supervised classification without the need for both class' labels compared to the state-of-the-art Deep SVDD and AnoGAN. 
\section{Datasets}
We used images from two different datasets. MNIST \cite{lecun-mnisthandwrittendigit-2010} dataset that contains 60,000 training images of handwritten digits and 10,000 test images. Fashion-MNIST \cite{xiao2017fashion} is a dataset of Zalando's article images—consisting of a training set of 60,000 examples and a test set of 10,000 examples.
All gray-scale images were resized to \(64 \times 64\) pixels.

\begin{figure}[h!]
  \centering
  \includegraphics[scale=0.3]{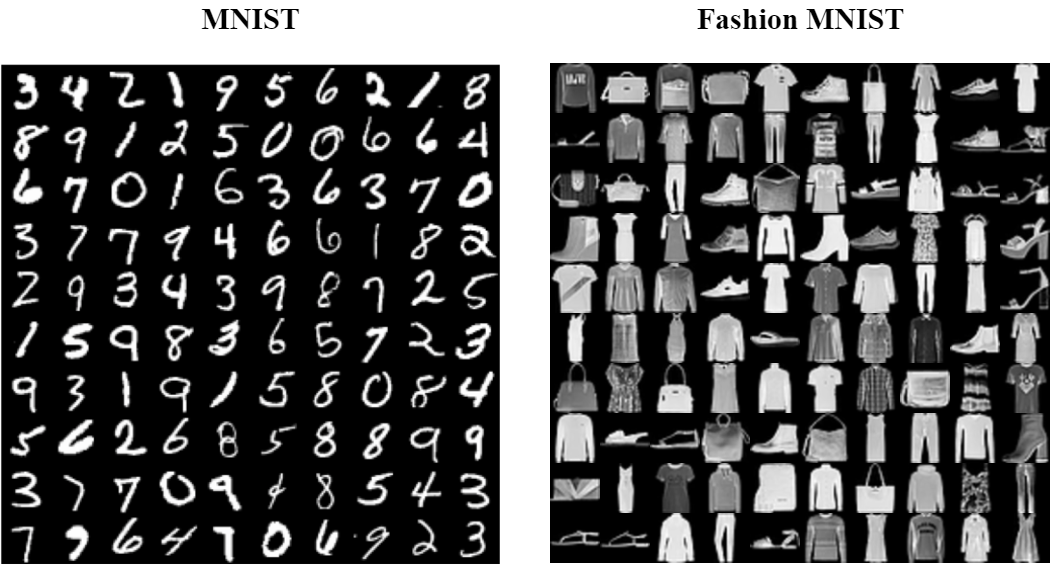}
  \caption{MNIST and Fashion MNIST sample images}
  \label{fig:DATA}
\end{figure}

\section{Generative Adversarial Networks}

A GAN is a deep learning model comprised of two main parts; Generator (\textbf{G}) and Discriminator (\textbf{D}). \textbf{G} can be seen as an art forger that tries to reproduce art-work and pass it as the original. \textbf{D}, on the other hand, acts as an art authentication expert that tries to tell apart real from forged art. Successful training of a GAN is a battle between \textbf{G} and \textbf{D} where if successful, \textbf{G} generates realistic images and \textbf{D} is not able to tell the difference between \textbf{G}'s generated images compared to real images.
\textbf{G} takes as input a random Gaussian noise vector and generates images through transposed convolution operations. \textbf{D} is trained to distinguish the real images \((x)\) from generated fake images \((G(z))\). Optimization of \textbf{D} and \textbf{G} can be thought of as the following game of minimax~\cite{goodfellow2014generative} with the value function $V(G, D)$: 
\begin{equation} \label{eq:1}
\min_G \max_D V(D, G) = \mathbb{E}_{x_{{\sim_P}_{data{(x)}}}} [\log D(x)] + \mathbb{E}_{z_{{\sim_P}_{z{(z)}}}} [\log (1 - D(G(z)))]
\end{equation}
During training, \textbf{G} is trained to minimize \textbf{D}'s ability to distinguish between real and generated images, while \textbf{D} is trying to maximize the probability of assigning "real" label to real training images and "fake" label to the generated images from \textbf{G}. The Generator improves at generating more realistic images while Discriminator gets better at correctly identifying between real and generated images. Today, when the term GAN is used, the Deep Convolution GAN (DCGAN) \cite{radford2015unsupervised} is the architecture that it refers to.

\subsection{Vanishing Twin GAN}
\begin{figure}[h!]
  \centering
  \includegraphics[scale=0.3]{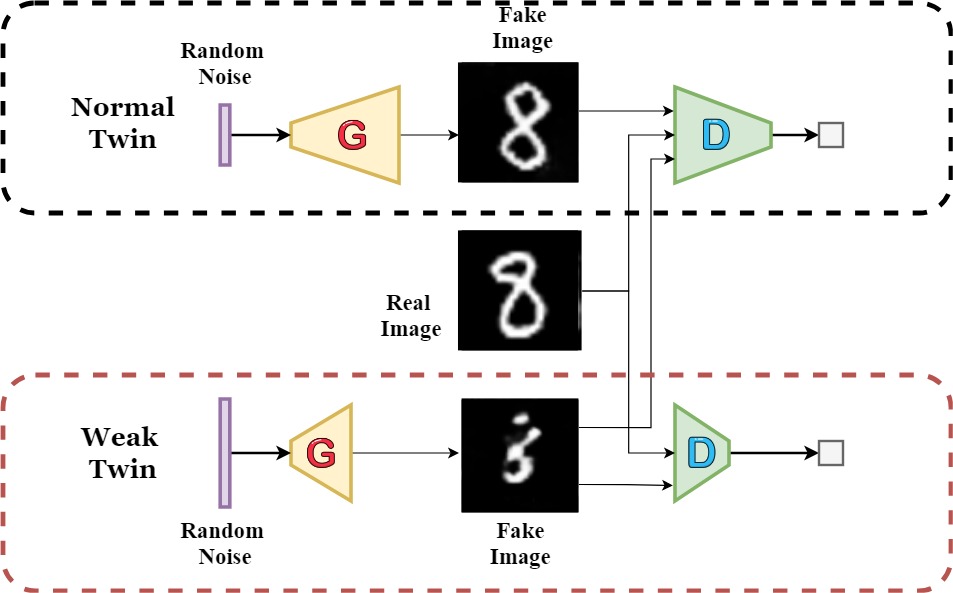}
  \caption{Vanishing Twin GAN architecture}
  \label{fig:twin}
\end{figure}

Vanishing Twin GAN sets to eliminate the need for labels for both classes that MCGAN relies on. Figure \ref{fig:twin} shows the architecture of Vanishing Twin GAN. The idea is to train two GANs in parallel. The \emph{Normal twin} which we want to train and use for classification of the images, and the \emph{Weak twin} which we want to use to improve the Normal twin's performance in image classification. By training a weak twin, our goal is to make the weak GAN's generator \textbf{G} get stuck in the noisy image generation stage that leads to Normal GAN's generalization problem. By training the weak twin, its generator does not fully represent the images of the class. However, the weak GAN should be good enough to learn to generate noisy versions of that class and not fall into mode collapse or not learning to generate images at all. Successful training of the weak GAN allows us to use its output as an input to the Normal twin's discriminator with a \emph{Fake} label. While MCGAN labeled real images from the similar class CII as \emph{Fake}, Vanishing Twin uses the same class's noisy, imperfect images with the \emph{Fake} label.

\subsubsection{Weakening the GAN}
We experimented with various modifications to the GAN's architecture to achieve our desired output. We found the following modifications to help train the weak GAN.

\begin{figure}[h!]
  \centering
  \includegraphics[scale=0.3]{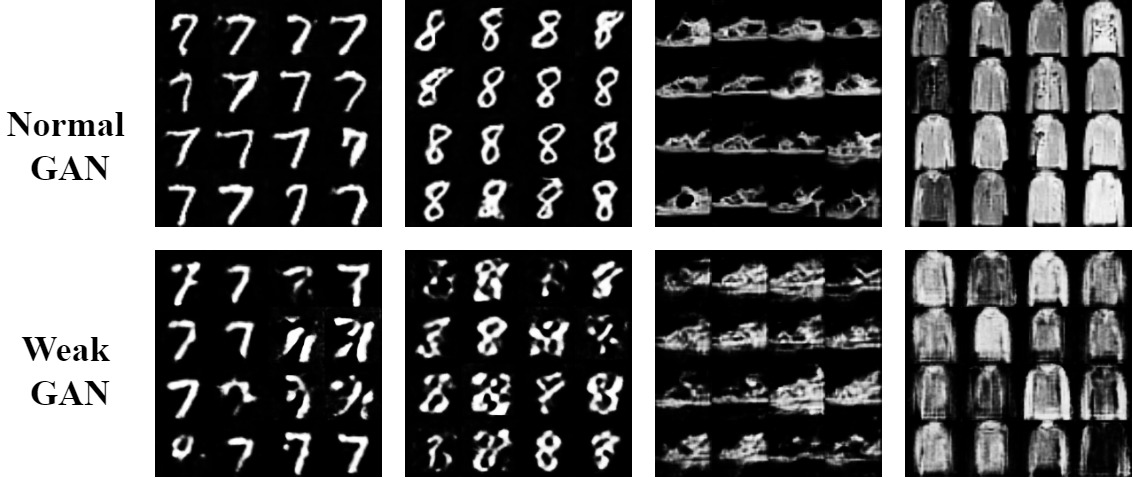}
  \caption{Regular and Weak GAN's generator outputs}
  \label{fig:weak}
\end{figure}

\begin{itemize}
  \item (i) \textbf{tuning the GAN's input noise dimension}     while decreasing the noise leads to mode collapse and the GAN not learning to generate new images, increasing the input noise dimension from a vector of size 100 (used in normal GAN) to 400 showed to be effective in hindering the learning of the weak GAN enough to make the outputs noisy.
  
  \item (ii) \textbf{making the network shallow}    while regular GAN's \textbf{G} has two layers of transposed convolution layers, we reduced the layers to one transposed convolution layer in the weak GAN. The discriminator \textbf{D}'s convolution layers also were reduced from two in regular GAN to one layer in weak GAN.
  
  \item (iii) \textbf{strides of the Transposed convolution and max-pool along with the height and width.} Increasing the strides of the transposed convolution and the immediate max-pooling layer of \textbf{G} proved to be effective in forcing the generator to generate noisy data.
\end{itemize}
Figure \ref{fig:weak} shows the regular and weak GANs' generator output during training. The GANs were trained to generate images of digits 7 and 8 from MNIST dataset and Sandals and Coats from Fashion MNIST dataset from left to right.

\subsection{Variation Score}
The Variation score $V(x)$ for the query image $x$, proposed by Schlegl \emph{et al.} \cite{schlegl2017unsupervised}, is defined as;
\begin{equation} \label{eq:ax}
V(x) = (1 - \lambda) \times \mathcal{L}_{R}({z}) + \lambda \times \mathcal{L}_{D}({z})
\end{equation}
where \(\mathcal{L}_{R}({z})\) (eq.~\ref{resi}) and \(\mathcal{L}_{D}({z})\) (eq.~\ref{disc}) are the residual  and discriminator loss respectively that enforce visual and image characteristic similarity between real image x and generated image \(G(z)\). The discriminator loss captures image characteristics using the output of an intermediate layer of the discriminator, \(f(.)\), making the discriminator act as an image encoder. Residual loss is the pixel-wise difference between image x and \(G(z\)).
\begin{equation}\label{resi}
\mathcal{L}_R({z}) = \sum|x - G(z)|
\end{equation}
\begin{equation}\label{disc}
\mathcal{L}_D({z}) = \sum|f(x) - f(G(z)|
\end{equation}
Before calculating V(x) in test, a point $z_i$ has to be found through back-propagation that tries to generate an image as similar as possible to image x. The loss function used to find $z_i$ is based on residual and discriminator loss defined below.
\begin{equation}\label{eq:4}
  \mathcal{L}({z_i}) = (1 - \lambda) \times \mathcal{L}_{R}({z_i}) + \lambda \times \mathcal{L}_{D}({z_i})
\end{equation}
$\lambda$ adjusts the weighted sum of the overall loss and variation score. We used $\lambda = 0.2$ to train our proposed MCGAN and AnoGAN~\cite{schlegl2017unsupervised}. Both architectures were trained with the same initial conditions for performance comparison.

\section{Competing Methods}
Ruff \emph{et. al} proposed a Deep One-class classification model (Deep SVDD) \cite{ruff2018deep} that outperformed shallow and deep semi-supervised anomaly detection models at the time, including AnoGAN. We compare our Vanishing Twin GAN against these models as baselines.
\subsection{Shallow Baselines}
We followed the same implementation details of the shallow models as used in Ruff \emph{et. al}'s Deep SVDD study.
(i) \textbf{One class SVM} (OC-SVM) \cite{scholkopf2001estimating} finds a maximum margin hyper-plane that best separates the mapped data from the origin. 
(ii) \textbf{Isolation Forest} \cite{liu2008isolation} (IF) isolates observations by randomly selecting a feature and then randomly selecting a split value between the maximum and minimum values of the selected feature. We set the number of trees to t = 100 and
the sub-sampling size to 256, as recommended in the
original work
\subsection{Deep Baselines}
Our Vanishing Twin GAN is compared with three deep approaches. (i) Ruff \emph{et. al}'s \textbf{Deep SVDD} showed improved accuracy of one class classification in a framework where one class from MNIST and CIFAR-10 \cite{krizhevsky2009learning} was kept as the known image, and the rest of the classes were treated as the anomaly. (ii) \textbf{AnoGAN} is trained as the base GAN benchmark for the task of image classification. (iii) We also trained a \textbf{NoiseGAN}, which, instead of using the generated images of a Weak GAN in the VTGAN, adds noise to the Real training image and feeds it to the discriminator with a \textit{Fake} label. We experimented with random Gaussian and Salt and Pepper noise. For AnoGAN, NoiseGAN, and VTGAN's Normal GAN, we fix the architecture to DCGAN \cite{radford2015unsupervised}.

\section{Experiments}
\begin{figure}[h!]
  \centering
  \includegraphics[scale=0.35]{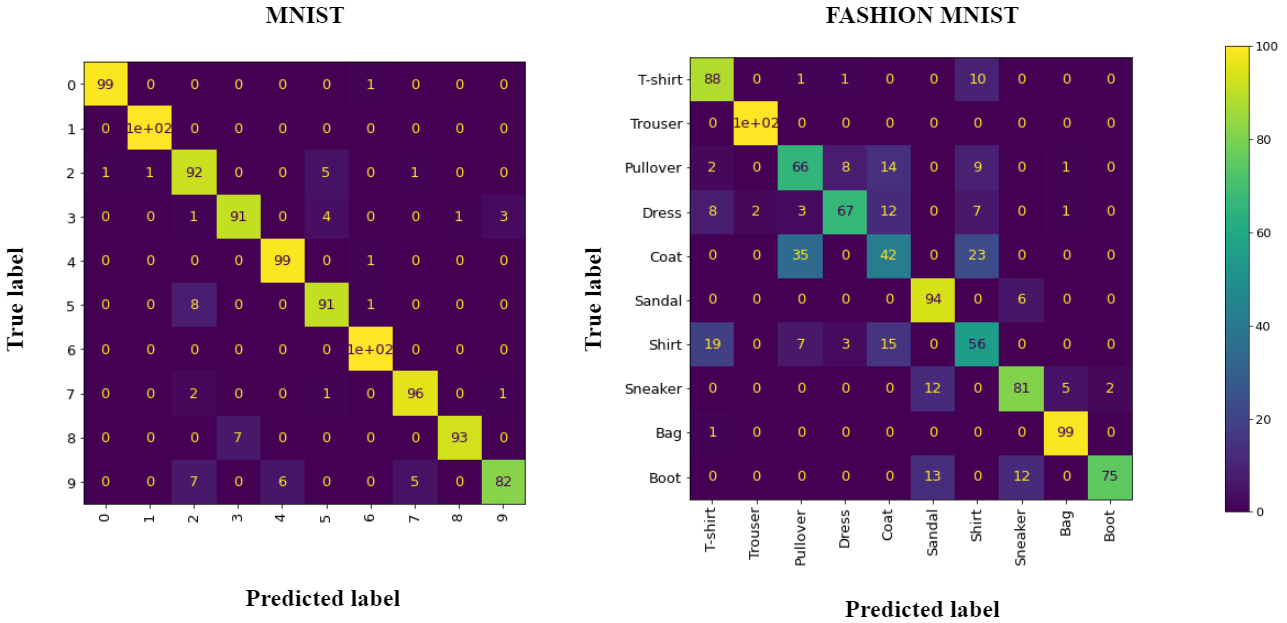}
  \caption{Confusion matrix of MNIST and F-MNIST embeddings}
  \label{fig:2}
\end{figure}

To pick a subset of similar classes from MNIST and Fashion-MNIST (F-MNIST) datasets that could cause generalization in GANs, we used metric learning \cite{kulis2012metric}. Metric learning aims to train models that can embed inputs into a high-dimensional space such that "similar" inputs are located close to each other. To bring images from the same class closer to each other via the embedding, the training data was constructed as randomly selected pairs of images from each class matched to the label of that class, instead of traditional \textit{(X,y)} pairs where y is the label for corresponding X as singular images of each class. By embedding the images using a shallow three-layer CNN, we computed the similarity between the image pairs by calculating the embeddings' cosine similarity. We used these similarities as logits for a softmax. This moves the pairs of images from the same class closer together. After the training was complete, we sampled 10 examples from each of the 10 classes and considered their near neighbors as a form of prediction; that is, does the example and its near neighbors share the same class. This is visualized as a confusion matrix shown in figure \ref{fig:2}. The numbers that lie on the diagonal represent the correct classifications, and the numbers off the diagonal represent the wrong labels that were misclassified as the true label. We intentionally used a shallow three-layer CNN to enforce some misclassification, as achieving near-perfect results in classifying datasets such as MNIST using CNNs is easy. Using the information from figure \ref{fig:2}, we picked the class pairs (9, 7) and (8, 3) from the MNIST dataset and (Coat, Shirt), (Coat, Pullover), and (Boot, Sandal) from F-MNIST dataset. 

\par For semi-supervised binary classification of the pair of a similar class of images, we trained a GAN on one class and used the variation scores for test images to classify the images for both classes. For each class pair (C1, C2), we experimented with once treating C2 as the unknown class and training GANs to generate images of class C1 and once treated C1 as the unknown class and trained GANs to generate images of C2. For instance, for the pair (9, 7), one AnoGAN / VTGAN / NoiseGAN was trained on 9s, and one was trained on 7s. For IF and OC-SVM, PCA was performed with 512 components, and the algorithms were executed on the mapped images. We followed Deep SVDD's implementation details and repeated the one-class classification for the (C1, C2) similar pairs.
The models were trained using an NVIDIA GeForce RTX 2080 Ti with 11 GB of memory.

\section{Results}

Table \ref{auc} shows the AUC for semi-supervised binary classification for each data pair for each model. Vanishing TWIN GAN outperforms AnoGAN in all training instances. Deep SVDD only outperforms Vanishing Twin GAN in 2 out of 10 instances.
It is interesting to see that shallow model (OC-SVM and IF) perform better than some of the deep models (AnoGAN, Deep SVDD)
\begin{table}[h!]
\begin{adjustbox}{width=\columnwidth,center}
 \begin{tabular}{c c c c c c c} 
 \hline
 
Known / Unknown & OC-SVM & IF & AnoGAN & NoiseGAN & Deep SVDD & VTGAN\\ [0.5ex] 
 \hline
 3 / 8 & 0.63& 0.64& 0.87& 0.88& 0.72& \textbf{0.90} \\
 8 / 3 & 0.59& 0.60& 0.90& 0.88& 0.90&\textbf{0.92}\\
 7 / 9 & 0.65& 0.64& 0.83& 0.83& 0.67&\textbf{0.85}\\
  9 / 7 & 0.75& 0.75& 0.83& \textbf{0.86} & 0.85& \textbf{0.86}\\
 \hline
     Boot / Sandal& 0.86& 0.85& 0.79& 0.80 & \textbf{0.98} & 0.87\\
    Sandal / Boot & 0.51& 0.51& 0.72& 0.74&0.54 & \textbf{0.76}\\
    Coat / Shirt & 0.55& 0.55& 0.68& 0.68& \textbf{0.76} & 0.70\\
    Shirt / Coat & 0.50& 0.50& 0.54& 0.56& 0.44 & \textbf{0.57}\\
    Pullover / Coat & 0.50& 0.50& 0.32& 0.61 & 0.45 & \textbf{0.62}\\
    Coat / Pullover & 0.54& 0.54& 0.67& 0.65& \textbf{0.70} & \textbf{0.70}\\
   \hline
\end{tabular}
\end{adjustbox}
\caption{One-class classification AUCs}
\label{auc}
\end{table}
Vanishing Twin GAN outperforms DCGAN in the task of image classification in all instances of classification between similar pairs of classes.

\begin{figure}[h!]
  \centering
  \includegraphics[scale=0.3]{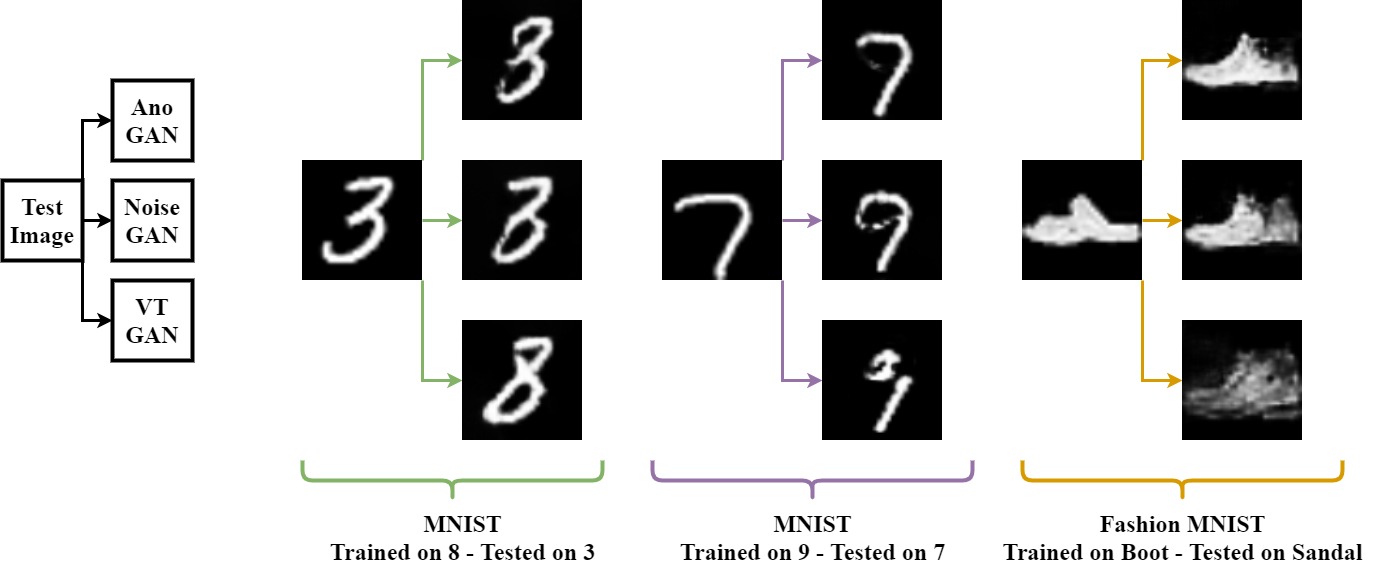}
  \caption{AnoGAN, NoiseGAN and VTGAN generated images}
  \label{fig:comp}
\end{figure}

Figure \ref{fig:comp} shows the visual differences of AnoGAN, NoiseGAN, and VTGAN on the same test image. Each test image is from the unknown class that GANs were not trained but are tested on. The first test image is an image of handwritten digit 3 from MNIST, while GANs trained on images of digit 8 generate a similar image to it. The next two test images are from MNIST, and Fashion MNIST, respectively, where  GANs trained to generate images of class 9 and Boot generate similar images to 7 and Sandal. We can see, in each instance, AnoGAN generated an image similar to the test image, even though it was trained on another class of images. Due to the similarity of the train and test classes, AnoGAN performs poorly in these instances for classification. VTGAN, on the other hand, generated images that were similar to the classes it was trained on, which translated to better performance for image classification. NoiseGAN also improved classification in some instances but underperformed compared to AnoGAN in others, showing adding only noise to images cannot implement the effect of VTGAN. 

\section{Conclusion}
We introduced Vanishing Twin GAN (VTGAN), a Deep model for semi-supervised image classification. Vanishing Twin GAN consists of two Generative Adversarial Networks, one of which (Weak twin) is designed for imperfect performance. Using the Weak twin's generated images, we improved the Normal twin's performance for semi-supervised image classification. VTGAN outperformed its GAN-based counterpart (AnoGAN) for anomaly detection in all test instances. VTGAN also outperformed the previous state-of-the-art Deep SVDD model for one-class classification of images in 7 out of 10 test instances, achieving the same performance in one instance and under-performed 2 instances. Our experiments showed the effect of VT training of a GAN by looking at the generated images from AnoGAN and VTGAN in settings where train and test data are drawn from similar classes of images (i.e. (3 / 8) and (Coat, Shirt)).

\section{Acknowledgements}
This research was funded by Chair in Medical Imaging and Artificial Intelligence funding, a joint Hospital-University Chair between the University of Toronto, The Hospital for Sick Children, and the SickKids Foundation.
\bibliography{references}

\end{document}